\documentclass[runningheads]{llncs}
\usepackage{pifont}
\usepackage{bm}
\usepackage{algorithm}
\usepackage{algorithmic}
\usepackage{graphicx}
\usepackage{amsmath} 
\usepackage{booktabs} 
\usepackage{subcaption} 
\begin{document}
\title{Bridging the Task Gap: Multi-Task Adversarial Transferability in CLIP and Its Derivatives}
%
%
\author{Kuanrong Liu\inst{1} \and
Siyuan Liang\inst{2} \and
Cheng Qian\inst{3}\and Ming Zhang\inst{3}\thanks{Corresponding author} \and
Xiaochun Cao\inst{1}}

\authorrunning{K. Liu et al.}

\institute{Shenzhen Campus of Sun Yat-sen University, China \and The National University of Singapore, Singapore \and
National Key Laboratory of Science and Technology on Information System Security, China.
}

\maketitle              
\begin{abstract}

As a general-purpose vision-language pretraining model, CLIP demonstrates strong generalization ability in image-text alignment tasks and has been widely adopted in downstream applications such as image classification and image-text retrieval. However, it struggles with fine-grained tasks such as object detection and semantic segmentation. While many variants aim to improve CLIP on these tasks, its robustness to adversarial perturbations remains underexplored. Understanding how adversarial examples transfer across tasks is key to assessing CLIP’s generalization limits and security risks.
 In this work, we conduct a systematic empirical analysis of the cross-task transfer behavior of CLIP-based models on image-text retrieval, object detection, and semantic segmentation under adversarial perturbations. We find that adversarial examples generated from fine-grained tasks (e.g., object detection and semantic segmentation) often exhibit stronger transfer potential than those from coarse-grained tasks, enabling more effective attacks against the original CLIP model.
Motivated by this observation, we propose a novel framework, Multi-Task Adversarial CLIP (MT-AdvCLIP), which introduces a task-aware feature aggregation loss and generates perturbations with enhanced cross-task generalization capability. This design strengthens the attack effectiveness of fine-grained task models on the shared CLIP backbone. Experimental results on multiple public datasets show that MT-AdvCLIP significantly improves the adversarial transfer success rate (The average attack success rate across multiple tasks is improved by over 39\%.) against various CLIP-derived models, without increasing the perturbation budget.
This study reveals the transfer mechanism of adversarial examples in multi-task CLIP models, offering new insights into multi-task robustness evaluation and adversarial example design.

\end{abstract}
\section{Introduction}



In recent years, Vision-Language Pretrained Models (VLPMs) have made significant progress in a variety of multimodal tasks~\cite{li2021albef}. CLIP~\cite{radford2021clip}, as a generalized vision-language alignment framework, has demonstrated excellent zero-shot generalization in tasks such as image-text retrieval and image classification~\cite{wang2019zeroclassification} through large-scale image-text contrastive pretraining. Meanwhile, CLIP has good task adaptability and has been widely extended to more challenging downstream tasks such as object detection and semantic segmentation after simple task-specific fine-tuning or module adaptation~\cite{gu2021VLIP,li2022glip,luo2023segclip,rao2022denseclip,zhou2023zegclip}.
CLIP still faces major challenges in fine-grained vision tasks. To improve its performance, methods like DenseCLIP and ZegCLIP adapt CLIP's multimodal representations for dense predictions. While these methods improve accuracy, their robustness—especially to adversarial attacks—and their relation to CLIP's original robustness remain underexplored. A key question in multi-task settings is whether adversarial examples~\cite{liang2020efficient,liang2022parallel,liang2022large,liu2024divide,kong2024patch,cheng2023topology,liu2019perceptual} from one task (e.g., detection or segmentation) can transfer to and affect other tasks or the base CLIP model. This cross-task vulnerability, termed multi-task adversarial robustness, is crucial for understanding representation transfer in multimodal models~\cite{liang2023badclip,chen2024less,chen2025less,liang2025revisiting} and for developing effective evaluation and defense strategies.
\begin{figure}[t]
    \centering
    \includegraphics[width=0.8\textwidth]{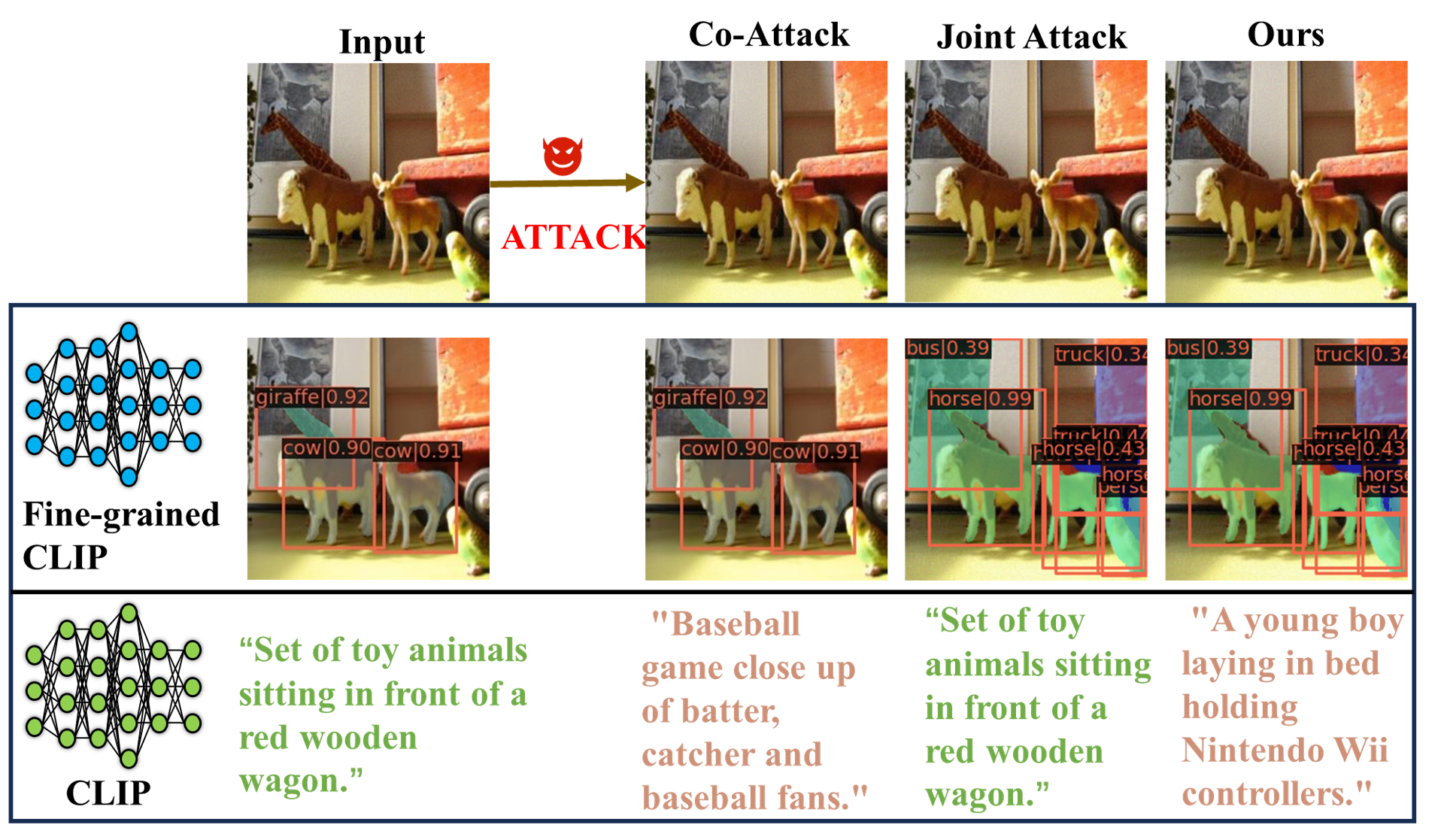} 
    \caption{Previous methods (e.g., Co-Attack) target only standard CLIP models and fail on fine-grained tasks. Even joint optimization ("Joint Attack" in the figure) mainly benefits fine-grained models. In contrast, our approach effectively attacks both fine-grained and standard CLIP models. The figure shows instance segmentation as a fine-grained task and image-text retrieval for CLIP. Red text marks successful attacks; green indicates correct outputs. Only our method generalizes well across both model types.}
    \label{fig:teaser}
\end{figure}
To tackle limited cross-task adversarial transferability~\cite{wang2023diversifying}, we propose MT-AdvCLIP, a multi-task adversarial attack framework for CLIP-based models. It employs a task-driven staged perturbation aggregation strategy with two core modules: Gradient-Guided Task Decoupling and Transferable Perturbation Weighting. First, strong perturbations are optimized on fine-grained tasks to exploit richer local gradients and avoid gradient conflicts. These are then refined under CLIP supervision to improve generalization in the global feature space. Perturbations from fine-grained models are weighted more heavily, while those directly on CLIP are scaled down to protect its feature space. Experiments across tasks (e.g., image-text retrieval, object detection on MSCOCO) show MT-AdvCLIP improves attack success by ~39
Extensive experiments demonstrate that our framework significantly enhances adversarial transferability across CLIP-based models. AS shown in Figure \ref{fig:teaser}, we evaluate our method on multiple tasks including image-text retrieval and object detection on the MSCOCO dataset, achieving approximately a 39\% improvement in average attack success rate over existing methods. 
Our main contributions are summarized as follows:
\begin{itemize}
    \item We conduct the first systematic analysis of adversarial transferability between the original CLIP model and its fine-grained task derivatives, highlighting the impact of gradient conflicts and the lack of structural information on robustness generalization.
    \item We propose a staged perturbation aggregation strategy that decouples task-specific gradients and enhances adversarial transferability across CLIP-based models.
    \item We validate our approach on multiple CLIP derivatives (such as DenseCLIP and ZegCLIP), where the average attack success rate reaches 89.15\% for DenseCLIP and 83.30\% for ZegCLIP, demonstrating strong robustness and generalization capability.
\end{itemize}

\section{Related Work}
\subsection{Vision-Language Pretrained Models}
Vision-Language Pretrained Models (VLPs) have advanced the development of multimodal understanding by jointly modeling the semantic alignment between images and texts~\cite{li2021albef}. Among them, CLIP (Contrastive Language–Image Pretraining)~\cite{radford2021clip} stands out as a representative approach. Trained on large-scale image-text pairs via contrastive learning~\cite{faghri2017vse++,chen2020simlr}, CLIP demonstrates impressive zero-shot generalization across various downstream tasks such as image classification~\cite{wang2019zeroclassification} and retrieval, without requiring task-specific fine-tuning. It employs dual encoders to project images and texts into a shared embedding space, where semantically aligned pairs exhibit higher similarity, enabling effective cross-modal understanding and reasoning. 
CLIP’s performance significantly declines on fine-grained tasks like object detection and semantic segmentation, highlighting challenges in generalizing to dense prediction. Prior works~\cite{zhong2022regionclip,rao2022denseclip,zhou2023zegclip} address this by improving region-level alignment: Zhong et al. introduce region-text alignment; Rao et al. reformulate matching at the pixel level; Zhou et al. design a lightweight decoder for zero-shot segmentation. However, the adversarial robustness of these models and their relation to the original CLIP remain underexplored. Our method provides the first systematic study on adversarial transferability between original CLIP and its task-specific variants, revealing key factors affecting cross-task robustness.

\subsection{Adversarial Attacks}
Adversarial attacks~\cite{costa2024adv,li2024adversarialsurvey} involve applying imperceptible perturbations to input data in order to mislead deep learning models into making incorrect predictions. This technique has been widely studied in image classification~\cite{goodfellow2014FGSM,madry2017PGD,carlini2017cwattack}, object detection~\cite{xie2017DAG,liu2018dpatch,liang2022GARSDC}, and semantic segmentation~\cite{rony2023proximalseg,maag2024uncertaintyseg}. Recent studies~\cite{zhou2023advclip,zhang2022coattack,lu2023SGA} reveal that CLIP and other VLPs, despite their strong zero-shot generalization, are notably vulnerable to adversarial examples. Due to their shared embedding space for both modalities, adversaries can manipulate image embeddings with minimal perturbations to push them closer to incorrect text embeddings, resulting in misleading predictions.

In addition, the multimodal nature of CLIP introduces new attack surfaces, such as disrupting the semantic alignment between image and text. Zhang \textit{et al.}~\cite{zhang2022coattack} first explored adversarial attacks on vision-language models by injecting perturbations into both modalities, demonstrating improved attack effectiveness. Zhou \textit{et al.}~\cite{zhou2023advclip}  generate universal adversarial perturbations for CLIP using generative adversarial networks (GANs). Lu \textit{et al.}~\cite{lu2023SGA}  extend single image-text pairs to set-level supervision across modalities, leveraging cross-modal consistency to generate transferable adversarial examples for VLPs. However, existing works primarily focus on CLIP’s behavior under standard tasks. The generalization of adversarial vulnerabilities from base CLIP to its fine-tuned or adapted variants for dense prediction tasks remains an open and underexplored area.However, current adversarial attacks primarily focus on single-task models, with limited research on generalized adversarial attacks across fine-grained and coarse-grained tasks. Due to the adversarial conflict between fine-grained and traditional tasks, existing methods struggle to achieve multi-task generalization. Our method effectively resolves gradient conflicts through segmented perturbation aggregation, enhancing the transferability of adversarial attacks across both fine-grained and coarse-grained tasks.
\begin{figure*}[h]
    \centering
    \includegraphics[width=0.9\textwidth]{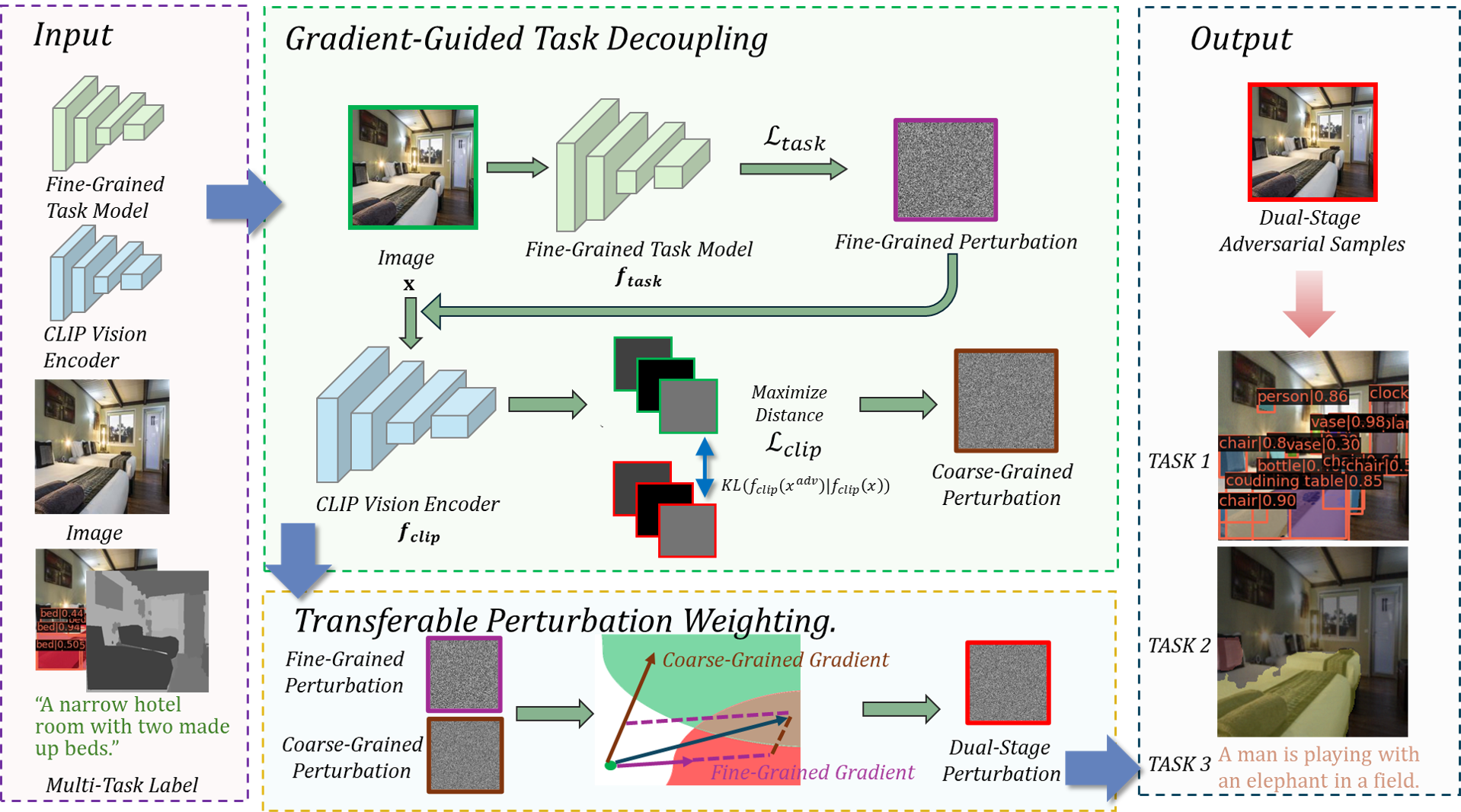} 
    \caption{The framework of our method consists of two stages. We first propose a two-stage perturbation stacking strategy to alleviate gradient conflicts between tasks, enabling the generation of generalizable adversarial examples. Then, we introduce a perturbation weighting mechanism to further enhance the generalization performance of the adversarial samples across multiple tasks. The final adversarial examples are capable of generalizing successfully to both the fine-grained CLIP-derived model and the original CLIP model.}
    \label{fig:framwork}
\end{figure*}

\section{Preliminaries}
\subsection{Threat Model}

We adopt DenseCLIP~\cite{rao2022denseclip} as the target model for object detection. DenseCLIP utilizes the pretrained CLIP model~\cite{radford2021clip} released by OpenAI as its backbone and fine-tunes it for detection tasks. We investigate the transferability of adversarial examples between the original CLIP model and DenseCLIP. Specifically, we use ResNet-50~\cite{he2016resnet} as the backbone architectures for CLIP and its derivatives. Additionally, we also explore the generalization of adversarial examples on semantic segmentation models, employing ZegCLIP~\cite{zhou2023zegclip} as the attack target, with ViT-B/16 as the backbone from the original CLIP.

\subsection{Attack Settings}

Although DenseCLIP and other derivative models leverage textual modality during training, they rely solely on the visual modality during inference. Therefore, to ensure the transferability of adversarial examples across different task models, we focus perturbations exclusively on the image modality, unlike prior CLIP adversarial attacks~\cite{zhang2022coattack,lu2023SGA} that simultaneously perturb image and text inputs.

We employ \textbf{Projected Gradient Descent (PGD)}~\cite{madry2017PGD} as the base untargeted attack method. PGD is an iterative first-order adversarial attack that generates adversarial examples by repeatedly applying gradient-based updates and projecting the perturbed inputs back into an $\ell_\infty$-bounded ball centered at the original input. Formally, given a clean image $x$, ground-truth label $y$, and loss function $\mathcal{L}$, the PGD update rule at step $t$ is defined as:

\begin{equation}
    {\mathbf{x}}^{t+1} = \Pi_{\mathcal{B}_\epsilon(\mathbf{x})} \left( \mathbf{x}^t + \alpha \cdot \text{sign} \left( \nabla_{\mathbf{x}} \mathcal{L}(f(\mathbf{x}^t), y) \right) \right),
\end{equation}
where $\Pi_{\mathcal{B}_\epsilon(\mathbf{x})}$ denotes the projection onto the $\ell_\infty$ ball of radius $\epsilon$ centered at $\mathbf{x}$, $\alpha$ is the step size, $f(\cdot)$ is the target model, and $\mathbf{x^0}$ is initialized as either the original image $\mathbf{x}$ or a randomly perturbed point within the allowed $\ell_\infty$ ball.



\section{Methodology}
\subsection{Observation}
For these derivative CLIP models, our goal is to generate adversarial examples that not only generalize to attack the models themselves but also retain their effectiveness against the original CLIP model. Intuitively, there are two possible approaches: first, since these derivative models are fine-tuned from CLIP, their image encoders may still preserve semantic information from the original CLIP. As a result, adversarial examples generated to attack these derivative models might also generalize to the original CLIP model. Second, inspired by multi-task optimization, one could combine the loss functions of both models during training to encourage the adversarial examples to generalize across both.


Our experiments show that neither of the two approaches achieves effective generalization. Although adversarial examples generated for derivative models can partially attack their source models, they are largely ineffective against the original CLIP, likely due to overfitting to task-specific features like bounding boxes in detection models. To address this, we attempted joint optimization using adversarial loss from both the original CLIP and the derivative model. However, this still failed to yield transferable examples—perturbations remained effective for detection but had little impact on CLIP's image-text retrieval task (see Section \ref{sec:exp} for details).
This suggests that the task-specific structural information introduced by additional components (e.g., decoders with bounding boxes) may bias the perturbation generation process, thus limiting the transferability of adversarial examples to the original CLIP. Furthermore, the representational gap caused by missing or altered task-specific features may help explain CLIP’s performance limitations on traditional vision tasks like detection and segmentation, despite its strong generalization elsewhere.

Additionally, we observe a gradient conflict during joint optimization: the adversarial gradients from the CLIP and derivative models often point in different directions, making it difficult to jointly optimize for both tasks. Interestingly, our findings suggest that the optimization direction of the derivative task tends to dominate. This is evidenced by the fact that adversarial examples generated through joint optimization retain attack effectiveness on the detection model, but fail to attack the original CLIP model effectively.
\subsection{Motivation and Overview}
To enable adversarial generalization across multiple CLIP-derived models and the original CLIP, we aim to design an optimization strategy that produces transferable adversarial examples. However, this is nontrivial due to the gradient conflicts introduced by different downstream tasks, which makes it difficult to directly obtain a unified perturbation direction.

Based on empirical observations, we identify two key properties:
\begin{itemize}
    \item \textbf{Task Dominance:} The optimization gradients from fine-grained tasks (e.g., detection or segmentation) tend to dominate during multi-task training.
    \item \textbf{Gradient Transferability:} Although adversarial examples crafted on downstream models show limited attack success on the original CLIP, the gradients from these models—being fine-tuned from CLIP—implicitly contain transferable directions that can generalize back to the original CLIP.
\end{itemize}

To leverage these insights, we propose a two-stage adversarial example generation framework that sequentially integrates task-specific and CLIP-guided optimization, enabling the perturbation to generalize effectively to the original CLIP.

\subsection{Stage I: Gradient-Guided Task Decoupling.}
We argue that fine-grained tasks exhibit stronger responses at the pixel level and in local features, thereby providing more precise and informative gradient signals for adversarial perturbations. These richer gradients enable faster convergence toward optimal directions and help avoid being trapped in local minima. In contrast, CLIP primarily focuses on global representations—such as high-level semantic concepts—when performing image-text matching. Due to the smoother and less distinctive nature of its gradients, the optimization process may be dominated by the fine-grained gradients during joint training. Therefore, we propose to decouple the two components and conduct attacks in a sequential manner to preserve the effectiveness of both.
In the first stage, we generate adversarial perturbations by attacking fine-grained task models such as object detection or semantic segmentation. The goal is to obtain a perturbation $\epsilon_{\text{task}}$ that is effective for the specific downstream task. Formally, we optimize:

\begin{equation}
\mathbf{\delta_{\text{task}}} = \arg\max_{\|\mathbf{\delta}\|_{\infty} \leq \epsilon_\text{task}} \mathcal{L}_{\text{task}}\left(f_{\text{task}}(\mathbf{x} + \mathbf{\delta}), y\right),
\label{eqa:loss_task}
\end{equation}

where $x$ is the clean input image, $y$ is the task label, $f_{\text{task}}$ denotes the downstream model (e.g., DenseCLIP or ZegCLIP),  $\mathcal{L}_{\text{task}}$ is the task-specific loss, $\mathbf{\delta}$ denote the adversarial perturbation, $\epsilon_\text{task}$ denotes the perturbation upper bound for the task. These perturbations often focus on pixel-level fine-grained features and dominate the optimization dynamics.
In the second stage, we refine the perturbation to enhance its generalization to the original CLIP model. Building on the task-specific perturbation $\epsilon_{\text{task}}$, we further optimize the perturbation using the CLIP image encoder $f_{\text{clip}}$. We adopt the Kullback–Leibler (KL) divergence to enforce deviation in CLIP’s output distribution over candidate text embeddings. 

Let $\mathbf{p} = f_{\text{clip}}(\mathbf{x})$ and $\mathbf{q} =f_{\text{clip}}(\mathbf{x} + \mathbf{\delta_{\text{task}}} + \delta) $ Then, the objective is:

\begin{equation}
\mathbf{\delta_{\text{clip}}} = \arg\max_{\|\mathbf{\delta}\|_{\infty} \leq \epsilon_{\text{clip}}}D_{\text{KL}} \left( \mathbf{p} \, \| \, \mathbf{q} \right),
\label{eqa:loss_clip}
\end{equation}
where $D_{\text{KL}}(\mathbf{p} \,\|\, \mathbf{q}) = \sum_i p_i \log \frac{p_i}{q_i}$ is the KL divergence between the clean and perturbed prediction distributions over text embeddings.
\subsection{Stage II: Transferable Perturbation Weighting.}
At this stage, we consider how to allocate perturbation weights to generate adversarial examples with stronger generalization capabilities. Since fine-grained tasks contain richer information, we assign them higher weights in the perturbation generation process. Moreover, during the training of fine-grained models, CLIP is used as a supervisory signal, which implicitly aligns the model with CLIP’s global semantics. As a result, although the generated adversarial examples may not explicitly optimize for CLIP, they inherently possess better generalization potential across both tasks, allowing faster convergence to a shared adversarial boundary. Additionally, since CLIP can be effectively attacked with relatively small perturbations, we allocate fewer resources to its perturbation component, thereby preserving the fine-grained perturbation structure and mitigating the interference introduced by CLIP perturbations. Specifically, the final adversarial perturbation and the upper bound of the perturbation can be formulated as:

\begin{align}
\label{eqa:delta}
&\mathbf{\delta} = \mathbf{\delta_{\text{task}}} + \mathbf{\delta_{\text{clip}}},\\
&\epsilon = \epsilon_\text{task}+\epsilon_\text{clip},\\
&\frac{\epsilon_\text{task}}{\epsilon_\text{clip}} = \lambda,
\label{lambda}
\end{align}

This formulation explicitly pushes the CLIP model’s output away from the clean prediction distribution, thereby increasing the transferability of adversarial examples across models. The perturbation for the fine-grained task ($\epsilon_{\text{task}}$) is set larger than that for CLIP ($\epsilon_{\text{clip}}$), i.e., $\lambda > 1$. 

Fig.~\ref{fig:framwork} shows the framework for MT-AdvCLIP, this staged strategy allows the first-phase perturbation to exploit task-sensitive features, while the second-phase refinement harnesses CLIP's feature structure to guide transferability. This framework effectively bridges the adversarial generalization gap between fine-grained task models and the original CLIP.
\begin{table}[ht]
\centering
\caption{Comparison with existing attack methods under multi-task adversarial attack. PGD refers to adversarial attacks on single-task models. Our method achieves high ASR across both tasks.
}
\resizebox{0.9\linewidth}{!}{ 
\begin{tabular}{lcccc}
\hline
\textbf{Method}          & \textbf{mAP $\downarrow$ (Detection)} & \textbf{Recall@1 $\downarrow$ (Retrieval)} & \textbf{ASR (mAP)} & \textbf{ASR (Recall@1)} \\ \hline
Clean & 24.0 & 46.24 & 0.0\% & 0.0\% \\
PGD (CLIP) & 20.3 & 0.1 & 15.8\% & 99.8\% \\
PGD (DenseCLIP) & 0.0 & 42.08 & 100.0\% & 9.0\% \\
Joint Optimization  & 0.0 & 39.1 & 100.0\% & 15.7\% \\
Dispersion Amplification~\cite{haleta2021multitask} & 0.02 & 42.32 & 99.9\% & 8.5\% \\
Co-ATTACK~\cite{zhang2022coattack} & 20.4 & 0.02 & 15.0\% & 99.9\% \\
SGA~\cite{lu2023SGA} & 18.8 & 0.04 & 21.7\% & 99.9\% \\
\textbf{MT-AdvCLIP} & 5.1 & 0.24 & 78.8\% & 99.5\% \\

\bottomrule
\label{tab:attack method}
\end{tabular}
}
\end{table}
\section{Experiment}
\label{sec:exp}
\paragraph{Downstream Tasks and Datasets}
To evaluate the effectiveness of our adversarial examples across diverse task models, we adopt MSCOCO~\cite{lin2014mscoco} as the common dataset for both training and evaluation. Our experiments span three tasks: image-text retrieval, object detection, and semantic segmentation. For the original CLIP model, we conduct experiments on the image-text retrieval task. For task-specific models, we consider DenseCLIP~\cite{rao2022denseclip} for object detection and ZegCLIP~\cite{zhou2023zegclip} for semantic segmentation. Notably, the segmentation model is trained on the COCO-Stuff-164K~\cite{caesar2018cocostuff} dataset, which augments MSCOCO with additional semantic categories. To ensure fair comparison across tasks, we restrict the segmentation evaluation to the original 80 object categories defined in MSCOCO.
\paragraph{Model Settings}
For the CLIP model, we use the pretrained weights released by OpenAI, which were trained on a dataset of 400 million image-text pairs and demonstrate strong multimodal representation capabilities. The model employs a Transformer-based text encoder, and we experiment with ResNet-50~\cite{he2016resnet} (RN50) and ViT-B/16~\cite{dosovitskiy2020vit} as vision encoders.

For fine-grained task models, we primarily use the DenseCLIP-based object detection framework in our experiments. The detection architectures is Mask R-CNN~\cite{he2017maskrcnn} with backbones of RN50.

We also evaluate the adversarial performance on the semantic segmentation model ZegCLIP~\cite{zhou2023zegclip}, which uses ViT-B/16 as the visual backbone. During adversarial attack experiments, we ensure that the CLIP model and the corresponding fine-grained task model share the same backbone architecture to maintain consistency.

\paragraph{Adversarial Attack}
All adversarial attacks are constrained within an $L_\infty$ perturbation budget of 8/255. For our method, we divide the perturbation between the task-specific model and the original CLIP model using a 3:1 ratio, assigning 6/255 to the fine-grained task model and 2/255 to the CLIP-guided refinement. This allocation reflects the dominant role of fine-grained features in generating transferable perturbations. For both our method and all baseline attacks, we set the step size to 2/255 and the number of iterations to 10 for each stage of the attack.
\paragraph{Evaluation Metrics}
To comprehensively assess the effectiveness of adversarial attacks across different downstream tasks, we adopt standard evaluation metrics specific to each task:

\begin{itemize}
    \item \textbf{Image-Text Retrieval:} For the image-to-text retrieval task on the original CLIP model, we report Recall@1 (R@1), which measures the percentage of times the correct text is ranked as the top-1 prediction given an image. A lower R@1 after attack indicates a more successful perturbation in misleading the model's retrieval output.
    
    \item \textbf{Object Detection:} For detection tasks using models like DenseCLIP, we report the mean Average Precision (mAP) following the MSCOCO evaluation protocol. The mAP metric reflects the model’s ability to correctly localize and classify objects. A drop in mAP under adversarial conditions indicates effective attack performance.
    
    \item \textbf{Semantic Segmentation:} For segmentation tasks using models such as ZegCLIP, we evaluate the mean Intersection-over-Union (mIoU) over all classes. mIoU measures the overlap between predicted segmentation maps and ground truth labels. A lower mIoU under attack suggests higher attack 
    success in distorting pixel-level predictions.
\end{itemize}
Attack Success Rate(ASR) is used to measure the effectiveness of an adversarial attack by comparing the model's performance before and after the attack. The formula is as follows:

\begin{equation}
   \text{ASR} = \frac{\text{ACC}_{\text{before}} - \text{ACC}_{\text{after}}}{\text{ACC}_{\text{before}}} 
   \label{eqa:ASR}
\end{equation}

 $ACC_\text{before}$ represents the model's accuracy on the original data (without any adversarial perturbations), and $ACC_\text{after}$ is the accuracy after adversarial examples are introduced, reflecting the performance degradation due to the attack. This ratio quantifies the relative decrease in accuracy, providing a standardized measure of attack success rate.
\begin{table*}[t]
\centering
\caption{Comparison of different perturbation orders. "A $\rightarrow$ B" means the attack is first applied to model A, then to model B. MT-AdvCLIP achieves the best performance.
}
\label{tab:order-ablation}
\resizebox{0.8\textwidth}{!}{
\begin{tabular}{lcccc}
\toprule
Perturbation Order & $\epsilon_{\text{fine}}$ & $\epsilon_{\text{clip}}$ & ASR (mAP) & ASR (Recall@1) \\
\midrule
Joint Optimization & - & - & 100.0\% & 0.0\% \\
CLIP $\rightarrow$ Fine-grained & 2/255 & 6/255 & 58.5\% & 98.6\% \\
CLIP$\rightarrow$ Fine-grained & 6/255 & 2/255 & 98.0\% & 14.5\% \\
Fine-grained $\rightarrow$ CLIP & 2/255 & 6/255 & 60.7\% & 99.4\% \\
Fine-grained $\rightarrow$ CLIP (MT-AdvCLIP) & 6/255 & 2/255 & \textbf{89.8\%} & \textbf{99.4\%} \\

\bottomrule
\label{tab:order}
\end{tabular}}
\end{table*}
\subsection{Comparison with Other Attack Methods}
In this section, we investigate the generalization performance of previous attack methods and our proposed approach on both fine-grained CLIP models and the original CLIP model. The evaluation includes adversarial attack methods for CLIP models~\cite{zhang2022coattack,lu2023SGA}, multi-task adversarial attack methods~\cite{haleta2021multitask}, and attack methods on fine-grained models. As shown in Table~\ref{tab:attack method}, the following conclusions can be drawn from the results:
\ding{182} Previous methods typically achieve good performance only on a single task. This is because they are designed with task-specific information in mind and do not incorporate signals from other tasks. This limitation is particularly evident in multimodal adversarial attacks such as Co-Attack and SGA, which also introduce adversarial noise in the text modality. Although they leverage the CLIP text encoder during training for certain fine-grained tasks, most evaluations rely solely on the visual modality, resulting in suboptimal performance on these tasks. In contrast, our method utilizes information from both tasks, enabling the generation of adversarial examples that generalize across multiple tasks.
\ding{183} In previous multi-task scenarios~\cite{haleta2021multitask,guo2024stealthy}, all tasks typically shared the same backbone. However, subtle differences in the weights and architecture between CLIP and its fine-tuned variants make previous multi-task attack methods less effective in this context. Moreover, directly applying joint multi-task optimization often leads to gradient conflicts, where gradients from fine-grained tasks dominate and suppress gradients for CLIP tasks, hindering generalization. Our approach addresses this issue by capturing the relationship between CLIP and fine-grained tasks. By establishing the dominance of fine-grained tasks during adversarial optimization, we construct composite adversarial examples and identify effective gradient directions for optimization.

\subsection{The Effect of Gradient Direction Dominated by Fine-Grained Tasks} 
In this section, we investigate the importance of applying perturbations from the fine-grained task first. We compare three strategies: applying CLIP perturbations before adding fine-grained perturbations, performing joint optimization, and our proposed method. As shown in Table~\ref{tab:order}, we draw the following conclusions: \ding{182} Later-added perturbations inevitably interfere with the effects of earlier ones, while perturbations derived from fine-grained tasks demonstrate higher robustness. Even when subsequent perturbations are added, fine-grained adversarial examples retain stronger attack performance. This is because fine-grained tasks inherently capture richer image information from their models, leading to more resilient adversarial samples. \ding{183} Both joint adversarial distribution and changes in perturbation order reduce the effectiveness of adversarial samples on the original CLIP model. This is attributed to gradient conflicts during the joint optimization of two tasks, where gradients tend to be minimized towards the fine-grained task. In contrast, our method identifies a feasible multi-task optimization route that produces adversarial examples effective on both models.

\subsection{Impact of Perturbation Magnitude and Composition}
In this section, we investigate how different perturbation magnitudes affect the performance of our method. Specifically, we vary the parameter $\lambda$ in Equation~\ref{lambda} to control the overall perturbation strength, setting it to 2, 4, 6, and 8. The corresponding experimental results are summarized in Table~\ref{tab:perturbation-ablation} (a) and Table~\ref{tab:perturbation-ablation} (b).
From the results, we derive the following observations and conclusions:
\ding{182} Our method successfully achieves adversarial generalization across both the fine-grained CLIP model and the original CLIP model. It substantially degrades the performance of object detection models and image-text retrieval tasks. This is attributed to the use of gradient information from both models during adversarial perturbation generation, enabling joint optimization.
\ding{183} A higher proportion of fine-grained adversarial noise within the overall perturbation leads to stronger attack performance. Although perturbation fusion may slightly reduce the effectiveness in individual tasks, the additional noise applied to the original CLIP model—characterized by localized fragmentary patterns—shows relatively stable impact. Even small perturbations can effectively disrupt CLIP’s performance. In contrast, maintaining robust attack effectiveness on pixel-sensitive tasks such as object detection requires a higher ratio of fine-grained information in the perturbation, which enhances the adversarial generalization across different tasks.


\begin{table}[h]
\centering
\caption{Effect of perturbation split on attack performance (DenseCLIP \& ZegCLIP).}
\label{tab:perturbation-ablation}
\setlength{\tabcolsep}{4pt}            
\renewcommand{\arraystretch}{0.95}

\begin{subtable}{0.48\linewidth}
\centering
\caption{DenseCLIP}
\resizebox{\linewidth}{!}{
\begin{tabular}{ccccc}
\toprule
Total $\epsilon$  & $\epsilon_{\text{fine}}$ & $\epsilon_{\text{clip}}$ & ASR (mAP)$\uparrow$ & ASR (R@1)$\uparrow$ \\
\midrule
4/255 & 2/255 & 2/255 & 15.4\% & 99.8\% \\
6/255 & 2/255 & 4/255 & 43.8\% & 97.4\% \\
6/255 & 4/255 & 2/255 & 64.2\% & 97.5\% \\
8/255 & 2/255 & 6/255 & 40.0\% & 99.4\% \\
8/255 & 4/255 & 4/255 & 45.8\% & 99.1\% \\
8/255 & 6/255 & 2/255 & 78.8\% & 99.5\% \\
\bottomrule
\end{tabular}}
\end{subtable}\hfill
\begin{subtable}{0.48\linewidth}
\centering
\caption{ZegCLIP}
\resizebox{\linewidth}{!}{
\begin{tabular}{ccccc}
\toprule
Total $\epsilon$ & $\epsilon_{\text{fine}}$ & $\epsilon_{\text{clip}}$ & ASR (mIoU)$\uparrow$ & ASR (R@1)$\uparrow$ \\
\midrule
4/255 & 2/255 & 2/255 & 52.5\% & 93.4\% \\
6/255 & 2/255 & 4/255 & 49.4\% & 99.3\% \\
6/255 & 4/255 & 2/255 & 65.9\% & 91.3\% \\
8/255 & 2/255 & 6/255 & 45.8\% & 99.3\% \\
8/255 & 4/255 & 4/255 & 62.9\% & 98.7\% \\
8/255 & 6/255 & 2/255 & 74.9\% & 91.7\% \\
\bottomrule
\end{tabular}}
\end{subtable}

\end{table}

\subsection{Visualization}

\begin{figure}[h]
    \centering
    \includegraphics[width=0.75\textwidth]{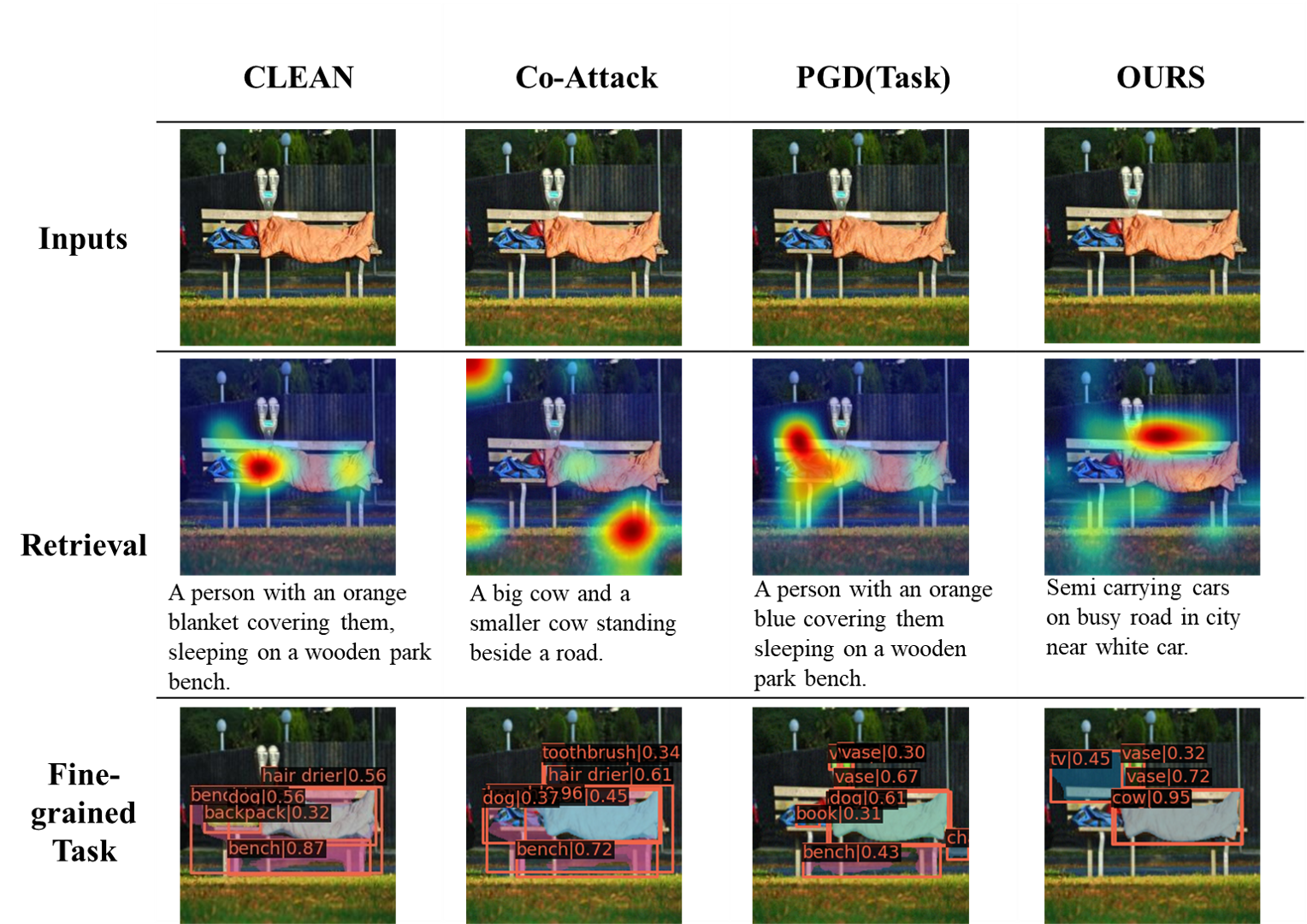} 
    \caption{Visualization of various adversarial examples generated by DenseCLIP. For the image-text matching task, the most relevant text along with the corresponding heatmap is shown, while the fine-grained task is illustrated using instance segmentation.}
    \label{fig:visualize}
\end{figure}
As shown in Fig.~\ref{fig:visualize}, we draw the following conclusion: the perturbations generated by our method can simultaneously mislead the model on both tasks. This is attributed to our optimization along a well-aligned gradient direction that enables effective multi-task adversarial perturbation generation.

\section{Conclusion}
In this paper, through systematic experiments and analysis, we reveal a significant generalization bottleneck in existing adversarial example generation strategies between the original CLIP model and its fine-tuned variants. Specifically, adversarial examples generated by optimizing task-specific models can effectively attack those models, but fail to generalize to the original CLIP. Meanwhile, joint loss optimization suffers from gradient conflicts across tasks, ultimately resulting in limited attack performance on CLIP. To address this issue, we propose a two-stage perturbation accumulation strategy to enable controllable enhancement and generalization of adversarial perturbations. Our method not only improves the attack effectiveness on the original CLIP model but also offers a novel perspective for resolving gradient conflicts and achieving cross-architecture transferability in multi-task adversarial attacks.

\section{Acknowledgements}
This research was funded by the Foundation of National Key Laboratory of Science and Technology on Information System Security (No. 6142111230501), Ningbo Science and Technology Innovation 2025 Major Project (2025Z027), National Natural Science Foundation of China (No.62025604, 62411540034), and the Fundamental Research Funds for the Central Universities, Sun Yat-sen University under Grants No. 23xkjc010. 

%
%
%
\bibliographystyle{splncs04}
\bibliography{mybib}

\end{document}